\documentclass[letterpaper,times]{IONconf}
 \usepackage{epstopdf}
 \usepackage{lipsum}
\usepackage[numbers]{natbib}
 \usepackage{color,soul}
\usepackage{booktabs}
\usepackage[utf8]{inputenc}
\usepackage{amsmath,amssymb,amsfonts}
\usepackage{xcolor}
\usepackage{hyperref}
\usepackage{algorithmic}
\usepackage{textcomp}
\usepackage{rotating}
\usepackage{multicol}
\usepackage{multirow}
\usepackage{multicol}
\usepackage{comment}
\usepackage{caption}
\usepackage{indentfirst}
\usepackage{subcaption}
\usepackage{authblk}
\makeatletter
\newcommand*{\rom}[1]{\expandafter\@slowromancap\romannumeral #1@}
\makeatother
\usepackage[flushleft]{threeparttable}

\DeclareMathOperator*{\argmin}{arg\,min}
\usepackage{sansmath}
\usepackage{comment}
\usepackage[english]{babel}
\usepackage{graphicx,float}
\usepackage{bm}
\usepackage{url}
\title{A Comparison of Robust Kalman Filters for Improving Wheel-Inertial Odometry in Planetary Rovers  
}

\author[1]{Shounak Das}
\author[1]{Cagri Kilic}
\author[2]{Ryan Watson}
\author[1]{Jason Gross}
\affil[1]{Department of Mechanical and Aerospace Engineering \\
West Virginia University \\
Morgantown, USA \\}
\affil[2]{The Johns Hopkins University Applied Physics Laboratory, Laurel, USA}

\begin{document}

\renewcommand{\sectionautorefname}{Section}
\renewcommand{\subsectionautorefname}{Section}
\renewcommand{\figureautorefname}{Fig.}
\renewcommand{\tableautorefname}{Table}
\renewcommand{\equationautorefname}{Eq.}

\maketitle

\begin{abstract}

This paper compares the performance of adaptive and robust Kalman filter algorithms in improving wheel-inertial odometry on low featured rough terrain. Approaches include classical adaptive and robust methods as well as variational methods, which are evaluated experimentally on a wheeled rover in terrain similar to what would be encountered in planetary exploration. Variational filters show improved solution accuracy compared to the classical adaptive filters and are able to handle erroneous wheel odometry measurements and keep good localization for longer distances without significant drift. We also show how varying the parameters affects localization performance. 
\end{abstract}


\section{Introduction}

One of the key problems in robotics is to estimate states like robot poses, locations of landmarks and sensor calibration parameters using the measurements from sensors. Due to inherent uncertainty in the measurements, system and measurement noise has to be taken into account, which are modelled using probability distributions. One of the most common noise distributions used in these problems is the Gaussian distribution \cite{thrun2002probabilistic}. Bayesian filtering\cite{barfoot2017state}\cite{sarkka2013bayesian} helps in estimating the posterior of the state given the measurements. For the linear Gaussian systems, this posterior has tractable analytical solution in the form of the well-known Kalman filter\cite{kalman1960new} which can also be extended to its nonlinear variants. But this assumption is not always valid and might lead wrong inference on the parameters. The Gaussian assumption leads to least-square problem which is highly affected by outliers. This is because of the unbounded nature of the squared error function.

In this paper, we modify the approach presented in  \cite{kilic2019improved} with robust filtering algorithms. The formulation in this paper is based on \cite{groves2015principles} and is an error-state Extended Kalman Filter~(EKF) which uses Inertial Measurement Unit~(IMU) measurements for error propagation and wheel odometry measurements as corrections. It also utilizes non-holonomic constraints and zero-velocity updates to reduce the impact of the IMU drift and wheel slippage. The advantage of this method is its low computational cost since it does not depend on cameras for visual odometry, which is suited for planetary rovers. It can also help in localization in places where visual odometry might fail, like shadow regions or places lacking visual features \cite{xue2019beyond}. By adding some of these existing methods, we extend upon the current approach to increase the robustness and accuracy of localization solution in presence of erroneous wheel odometry measurements. This paper offers the following contributions 1) it reviews several state of the art robust Kalman filters side-by-side within the same context 2) it offers an experimental evaluation of the methods discussed using a relevant platform and publicly available data-sets 3) the accompanying software implementations are made available in GitHub\footnote{\url{https://github.com/wvu-navLab/corenav-GP/tree/Robust_Methods}}. 

The rest of the paper is divided into six sections. Section \rom{2} reviews relevant literature on robust filtering. The base algorithm is discussed in the next section followed by a summarised theoretical discussion of the methods which are evaluated in section \rom{4}.  Sections \rom{5} and \rom{6} contain description of the experimental setup and software implementations. The results are presented and analyzed in section \rom{7} and concluded with section \rom{8}.

\section{Related work}\label{relatedwork}
\begin{figure}[htb!]
    \centering
    \includegraphics[width=0.5\columnwidth]{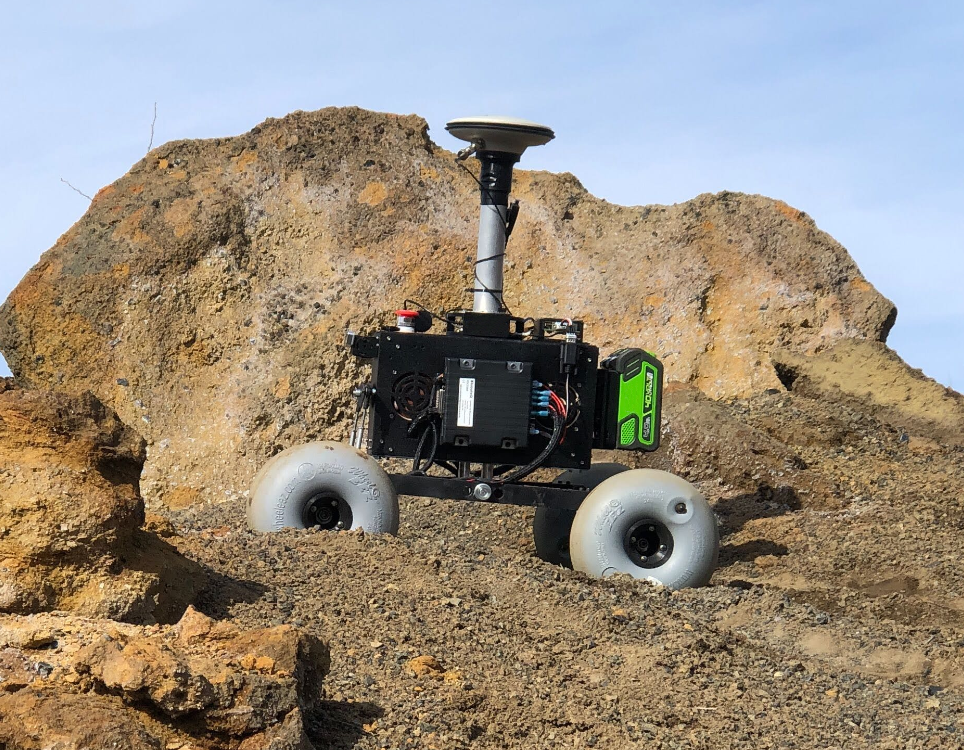}
    \caption{Pathfinder testbed platform}
    \label{fig:pathfinder}
\end{figure}

There is a rich body of literature that has considered modifying the standard Kalman filter to accommodate outliers. A covariance scaling method is introduced in \cite{chang2014robust} using the ratio between the Mahalonobis distance between the observed data and the predicted data and the critical value of $\chi^{2}$ distribution to reduce the effect of observation outliers or heavy-tailed measurement noise distribution. An adaptive factor is defined in \cite{yang2001adaptively} that helps the filter to balance the contribution of discrepancies due to dynamic modelling errors and new measurement as well as measurement outliers. \cite{fang2018robustifying} uses an adaptive innovation-saturation mechanism to estimate the changing measurement error bound and saturate the innovation process. An adaptive version of the Kalman filter is derived in \cite{akhlaghi2017adaptive} using a fading factor and the current innovation and residual values to recursively calculate process and measurement noise matrices.  The Huber cost function has been incorporated into the update step of the Kalman filter using M-estimation principles in \cite{durovic1999robust}\cite{karlgaard2015nonlinear}. Similar method has been used with tightly and loosely coupled GNSS/INS EKF with simulated faults in \cite{crespillo2018tightly}, where the robust version has comparable performance to that of fault detection and exclusion techniques. \cite{wang2019outlier} derives a robust filter using the Double-Gaussian Mixture Correntropy Loss between the data and the predicted measurements. Variational inference has also been used to approximate state posteriors with heavy tailed distribution for the measurement models in \cite{agamennoni2011outlier,agamennoni2012approximate,sarkka2013non,piche2012recursive,huang2017novel}. Faulty measurements are excluded using indicator variables with beta-Bernoulli prior and approximated with the state using mean-field variational inference in \cite{wang2018robust} in a simulated target tracking scenario. Bayesian weights are assigned to each measurement and estimated along with the filter parameters using Expectation-Maximization (EM) framework in \cite{ting2007kalman}. 

\section{Theory and equations}\label{probstatement}

In this section, we describe the Kalman filter algorithm and $CoreNav$ \cite{kilic2019improved, kilic2021slip} which we use as our base algorithm. The Kalman filter is the most well-known Bayesian filtering algorithm. It is a maximum likelihood estimator for the linear Gaussian model. The filter equations are 
\begin{equation}
   \hat{x}_{k/ k-1} =F_k \hat{x}_{k-1 / k-1} \label{eq:3} 
\end{equation}
\begin{equation}
  P_{k / k-1} =F_k P_{k-1 / k-1} F_k^{T}+Q_k  \label{eq:4}  
\end{equation}
\begin{equation}
  \hat{v}_k =z_k-H_k \hat{x}_{k / k-1} \label{eq:5}  
\end{equation}
\begin{equation}
  K_k =P_{k / k-1} H^{T}_k\left[H_k P_{k / k-1} H^{T}_k\right.+R_k]^{-1} \\\label{eq:6}  
\end{equation}
\begin{equation}
  \hat{x}_{k / k} =\hat{x}_{k / k-1}+K_k \hat{v}_k  \label{eq:7}  
\end{equation}
\begin{equation}
  P_{k / k} =[I-K_k H_k] P_{k / k-1} ,  \label{eq:8}  
\end{equation}

\noindent where $\hat{x}_{k / k}$ is the current estimate of the state using the measurement $z_k$. $F_k$, $H_k$ are the state transition and measurement matrices respectively. Matrix $P_{k/k}$ represents the state uncertainty,  $Q_k $ is the process noise covariance matrix and  $R_k$ is the measurement noise covariance matrix. The filter gain at time $k$ is denoted by $K_k$. 

As detailed in \cite{kilic2019improved}, $CoreNav$ is an error state EKF with the state vector represented in the local navigation frame by

\begin{equation}
    \mathbf{x}_{err}^{n} = \bigg
    (  \delta\mathbf{\Phi}_{nb}^{n}  \quad \delta \mathbf{v}_{eb}^{n} \quad \delta\mathbf{p}_{b} \quad \mathbf{b}_{a} \quad \mathbf{b}_{g} \bigg )^{
    \mathbf{T} } , \label{eq:1}
\end{equation}

where $\delta\mathbf{\Phi}_{nb}^{n}$ is the error in the attitude of the rover from the body frame ($b$) to the local navigation frame ($n$) expressed in the local navigation frame, $\delta \mathbf{v}_{eb}^{n}$ is the velocity error in the navigation frame, $\delta\mathbf{p}_{b}$ is the position error. $\mathbf{b}_{a}$ and $\mathbf{b}_{g}$ are IMU acceleration and gyro biases respectively. The position error vector is 

\begin{equation}
  \delta\mathbf{p}_{b} = \bigg (\delta\mathbf{lat}_{b} \quad \delta\mathbf{lon}_{b} \quad \delta\mathbf{h}_{b} \bigg )^{
    \mathbf{T}}.  \label{eq:2}
\end{equation}

which is expressed in latitude, longitude and height frame.
The states are propagated forward with measurements from IMU and corrected with wheel odometry measurements. The measurement $z_k$ is also an error vector represented as 

\begin{equation}
\delta \mathbf{z}_{O}=\begin{pmatrix}
{\tilde{v}}_{lon,O}- {\tilde{v}}_{lon,i}\\ -{\tilde{v}}_{lat,i}\\-{\tilde{v}}_{ver,i}\\ 
\tilde{\dot{{\psi}}}_{nb,o}-\tilde{\dot{{\psi}}}_{nb,i}\overline{cos{\hat{\theta}_{nb}}} 
\end{pmatrix} \,  \label{eq:2a}
\end{equation}

where $\tilde{\dot{{\psi}}}_{nb}$ and $cos{\hat{\theta}_{nb}}$ are heading rate and pitch angle of the rover body frame with respect to the navigation frame respectively. $\tilde{v}_{lon}$, $\tilde{v}_{lat}$, and $\tilde{v}_{ver}$ are predicted longitudinal, lateral and vertical rear wheel speed, respectively. The subscript $i$ is used for values derived from INS solution, and $O$ is used for values derived from the wheel odometry measurements. Motion constraints, such as zero velocity of the vehicle along the rotation axis of the wheels, and zero velocity perpendicular to the traversal surface are used as pseudo measurements to update the state vector \cite{dissanayake2001aiding}. It also utilizes zero-velocity-constraints \cite{skog2010zero}, i.e detects it has stopped using wheel velocity data and uses that information to maintain Inertial Navigation System (INS) alignment and reduce the rate of error growth. 


\section{Robust and Adaptive filters}

In this section, we provide a brief description of the robust and adaptive methods that have been evaluated in this paper. Acronyms have been defined (inside parenthesis) for ease of presenting the experimental results. 

\subsection{Huber Regression Kalman Filter (HKF)} 

M-estimators \cite{maronna2019robust,huber2004robust} are maximum-likelihood estimators for certain specific error distributions that differ from standard Gaussian distribution. They have been implemented successfully in many robotics applications due to lower sensitivity and high breakdown in presence of outliers \cite{bosse2016robust}. An iterative version of M-estimation has been discussed in \cite{karlgaard2015nonlinear,durovic1999robust} which uses the Huber cost function \cite{huber1992robust} at every time step instead of the squared error function. This causes the error between the predicted measurement and the actual measurement to grow like the squared function when below a parameter $\Delta$ and linearly when above it. 

The following derivation follows \cite{karlgaard2010robust}. A single filter prediction and update cycle can be written as a least squares problem 
 
\begin{equation}
 \begin{bmatrix}\hat{x}_{k/k-1}\\z_k\\\end{bmatrix} = \begin{bmatrix}I\\ H_k\\\end{bmatrix} x_k + E_k \label{eq:9}
\end{equation}
where 
\[E_k = \begin{bmatrix}\delta_k\\v_k\end{bmatrix}\]

and
\[\textbf{E}[E_kE_k^T] =\begin{bmatrix}P_{k/k-1} & 0\\ 0 & R_k\end{bmatrix} = S_kS_k^T . \]

Now the new linear equation can be written as
\begin{equation}
 Y_k = B_kx_k + \xi_k   \label{eq:10}
\end{equation}
where 
\[B_k =  S_k^{-1}\begin{bmatrix}I\\ H_k\\\end{bmatrix}\] 
\[\xi_k = -S_k^{-1}E_k\]
\[Y_k = S_k^{-1}\begin{bmatrix}\hat{x}_{k/k-1}\\ z_k\end{bmatrix}\]
\[\textbf{E}[\xi_k\xi_k^T] = I  \, . \]

Least squares solution of the modified linear equation is 

\[ \hat{x}_k = (B_k^TB_k)^{-1}B_k^TY_k\]
\[P_{k/k} = (B_k^TB_k)^{-1} \, . \]

\noindent\(S_k\) can be calculated by Cholesky decomposition of the covariance matrix of \(E_k\).

However, least squares is highly effected by outliers due to unbounded nature of the squared error function. So, a different cost function can be used  
\begin{equation}
 \hat{x}_{k/k} = \argmin_x J(x)  = \argmin_x\, \sum ^m_{i = 1} \rho\,( y_{ik} - b_{ik}x) , \label{eq:11}  
\end{equation}
\vspace{10 pt}

\noindent where \(m\) is the dimension of \(Y_k\), \(y_{ik}\) is the \(i^{th}\) element of \(Y_k\) and \(b_{ik}\) is the \(i^{th}\) row of \(B_k\). If \(\rho\) is quadratic, then it reduces to the least squares and the Kalman filter solution. But this function can be set such that the effect of large residuals is reduced. The algorithm is optimal for the Gaussian case. Thus, the function \(\rho\) can be squared for smaller values of the argument, but should increase slowly for larger values. Another important requirement is that the influence function \(\psi \) be continuous and bounded. It should be bounded so that no single measurement can affect the total cost largely and continuity helps in reducing rounding errors. One such function is the Huber cost function \cite{huber1992robust}
\begin{equation}
    \rho(z)=\begin{cases} z^{2} / 2 & |z| \leq \Delta \\ \Delta|z|-\Delta^{2} / 2 & |z|>\Delta \end{cases} \label{eq:12}
\end{equation}

\begin{equation}
    \rho^{\prime}(z)=\begin{cases} z & |z| \leq \Delta \\ \Delta \operatorname{sgn}(z) & |z|>\Delta \end{cases}  \label{eq:13}
\end{equation}

\begin{equation}
    \psi(z) = \frac{\rho{'}(z)}{z} \, . \label{eq:14}
\end{equation}


\noindent Other robust functions are discussed in \cite{bosse2016robust}. Let \(p\) be the dimension of \(x_k\). To minimize the cost function in (\ref{eq:11}), differentiate \(J\) and set it to zero

\begin{equation}
     \sum ^m_{i = 1} b_{ij}\psi\,( y_{ik} - b_{ik}\hat{x}_{k/k}) = 0  \qquad \qquad \text{for} \;j = 1\; ... \;p \, .    \label{eq:15}
\end{equation}

Assume \(\Psi = diag(\psi\,( y_{ik} - b_{ik}\hat{x}_{k/k}))\), which is a \(m\) dimensional diagonal matrix. Then (9) can be written as 

\[ B_k^T\Psi(B_k\hat{x}_{k/k} - Y_k) = 0  \, . \]


\noindent This equation cannot be solved explicitly since \(\Psi\) depends on the unknown state \(x_k\). However \(x_k\) can  be solved in an iterative manner using iterative re-weighted least squares 

\begin{equation}
    \hat{x}_{k/k}^{j+1} =  (B_k^T\Psi^{(j)}B_k)^{-1}B_k^T\Psi^{(j)} Y_k \, . \label{eq:16}
\end{equation}



\noindent This solution converges if \ \(\psi\) \  is non-increasing. The iteration can be started with the least-square solution \(\hat{x}_{k/k}^{0} =  (B_k^{T}B_k)^{-1}B_k^{T}Y_k\). After the state has converged, the covariance can be estimated by the expression 

\begin{equation}
    P_{k/k} = (B_k^T\Psi B_k)^{-1} \, . \label{eq:17}
\end{equation}

Choosing the tuning parameter \(\Delta\) is important, and depends on the perturbing parameter \(\epsilon\), which represents the proportion of contamination in the assumed residual distribution \cite{karlgaard2010robust}. If the perturbing parameter is known, then the optimal choice \(\Delta^*\) is given in \cite{huber2004robust}. When the perturbing parameter is unknown, \(\Delta\) is usually chosen between \(1.0\) and \(2.0\). 


\subsection{Covariance Scaling Kalman Filter(CSKF)}

The Mahalonobis distance ratio defined in equation (\ref{eq:19}) can be used for assessing if the measurement is inlier or outlier. Following \cite{chang2014robust}, the measurement noise covariance matrix or the measurement prediction covariance is scaled for every outlier detected using \(\chi ^2\) criterion. 

\begin{equation}
    M_k^2 = e_k^T[H_k P_{k / k-1} H^{T}_k + R_k]^{-1}e_k \, . \label{eq:18}
\end{equation}


\noindent\(M_k\) is the Mahalonobis distance of the actual measurement from the predicted measurement. Assuming that the measurement follows a Gaussian distribution, the test statistic that can be used to test if the measurement is an outlier is the critical value of the \(\chi^2\) distribution corresponding to significance level \(\alpha\) with degree of freedom equal to the measurement dimension, \(\chi_{m\alpha}\). This is also used as the \textit{Independent Innovation Test} in \cite{tong2011batch}.

\begin{equation}
    \gamma_k = \frac{M_k^2}{\chi_{m\alpha}} \label{eq:19}
\end{equation}


\noindent The scaling factor \(\gamma_k\) is used to modify the measurement noise covariance matrix \(R_k\). If \(\gamma_k\) is less than \(1\) then measurement \(Y_k\) is an inlier and \(R_k\) remains the same. But if scaling factor is greater than \(1\), denoting an outlier, the inflated \(R_k\) is given by \cite{chang2014robust}

\begin{equation}
    \hat{R}_k = (\gamma_k - 1)H_kP_{k / k-1} H^{T}_k + \gamma_k R_k \, . \label{eq:20}
\end{equation}


\noindent The scaling of $R_k$ results in de-weighting the suspected outlier measurements. The rest of the steps remain same as the standard filter equations,i.e replace $\hat{R_k}$ in place of $R_k$ in (\ref{eq:6}), (\ref{eq:7}), (\ref{eq:8}).

\subsection{ Variational Filters} 

Variational Bayes filters \cite{sarkka2009recursive} use a prior for auxiliary variable in the measurement model such that the marginalized measurement distribution has heavier tails. Due to the non-Gaussian nature of this model, state posteriors have to be approximated using variational inference \cite{murphy2012machine}. The state and the auxiliary variables are estimated iteratively until convergence. The most common method in variational inference uses the mean-field assumption to factor the joint distribution of the state and the auxiliary variable as a product of each of the variable distributions. Then, the posterior of both state and auxiliary variable are approximated by minimizing the Kullback–Leibler divergence \cite{mackay2003information} between the posterior and the product (i.e., maximizing the lower bound).

Here we have implemented the variational inference filters presented in \cite{agamennoni2012approximate}, \cite{sarkka2013non} and\cite{huang2017novel}. All three methods share the same inference method, they only differ in the auxiliary random variable that is considered in the measurement model. \cite{agamennoni2012approximate} assumes $R_k$ has a inverse-Wishart \cite{murphy2012machine} prior

\begin{equation}
    R_k \sim \mathcal{W}^{-1}(sR, s) \, , \label{eq:23}
\end{equation}


\noindent where $R$ is the nominal measurement noise, $ s > d-1 $ is the degree of freedom, $d$  the dimension of the measurements. The measurement model is Gaussian,

\begin{equation}
    z_k|x_k,R_k \sim \:N(H x_k,R_k) \, . \label{eq:24}
\end{equation}


\noindent Variational inference approximates the posterior of the state $q(x_k)$ with the standard filter correction step i.e Gaussian and $q(R_k)$ as 
\begin{equation}
    R_k |z_k \sim \mathcal{W}^{-1}( (s+1)\Lambda_k,s+1)  ,\label{eq:25}
\end{equation}
where 
\begin{equation*}
    \Lambda_k = \frac{sR + S_k}{s+1}
\end{equation*}
\begin{equation*}
    S_k = (z_k-H_k \hat{x}_{k / k-1})(z_k-H_k \hat{x}_{k / k-1})^{T} + H_k P_{k / k-1} H^{T}_k.
\end{equation*}


\noindent This posterior from \cite{agamennoni2012approximate} has an elegant interpretation. The posterior noise matrix is a convex combination of the nominal noise matrix and the expected sufficient statistics, $s$ denoting the relative importance between the two. 

\cite{sarkka2013non} assumes $R_k$ is fixed but scaled by a scalar random variable which follows a Gamma distribution \cite{walck2007hand}
\begin{equation}
    \lambda_k \sim Gamma(\frac{\nu}{2},\frac{\nu}{2}) \label{eq:26}
\end{equation}
and the measurement distribution is

\begin{equation}
    z_k|x_k,\lambda_k \sim \:N(H x_k,\frac{1}{\lambda_k}R_k) . \label{eq:27}
\end{equation}


\noindent Similar to the previous method, variational inference approximates $q(x_k)$ with the standard filter correction step and $q(\lambda_k)$ as 
\begin{equation}
    \lambda_k|z_k \sim Gamma(\frac{\nu + d}{2}, \frac{\nu + \Bar{\gamma}_k}{2})    \label{eq:28}
\end{equation}
where 
\begin{equation}
\begin{aligned}
    \Tilde{\gamma_k} &= \mathbf{E}_x[(z_k - h(x_k))^T R_k^{-1} (z_k - h(x_k))] \\
     &= \text{tr}(\mathbf{E}_x[(z_k - h(x_k))^T (z_k - h(x_k))]R_k^{-1}).
    \end{aligned} \label{eq:29}
\end{equation}

\noindent The expectation inside the $trace$ operator can be approximated using sigma-point operations \cite{van2004sigma}. The posterior in equation (\ref{eq:28}) makes sense when examining the mean which can be  written as 

\begin{equation*}
    \mathbf{E}(\lambda_k) = \frac{\nu + d}{\nu + \Bar{\gamma}_k} .
\end{equation*}

For the case of inliers, the measurements will be closer to their corresponding predictions resulting in smaller $\Tilde{\gamma_k}$, thus larger $\lambda_k$ and thus smaller uncertainty(larger weights). The opposite would happen for outliers resulting in larger $\Tilde{\gamma_k}$, thus smaller $\lambda_k$ and thus larger uncertainty(smaller weights).

The common aspect of both these methods is that after marginalizing out the auxiliary variable, the resulting measurement distribution has heavier tails (in this case, it is the Student's t  distribution \cite{murphy2012machine}). Both $s$ in \cite{agamennoni2012approximate} and $\nu$ in \cite{sarkka2013non} represent the extent of heavy-tailedness of the distribution, with $s,\nu = 1$ being the Cauchy distribution \cite{walck2007hand} and $s, \nu \rightarrow \infty$ denoting the Gaussian distribution. \cite{huang2017novel} assumes inverse-Wishart priors for both state uncertainty matrix and the measurement noise covariance matrix, which is assumed to be slowly drifting and following the same inference method as above. Readers are referred to \cite{huang2017novel} for further details. In the results sections, \cite{agamennoni2012approximate} is represented as $ORKF1$, \cite{sarkka2013non} as $ORKF2$ and \cite{huang2017novel} as $ORKF3$ ($ORKF$ stands for Outlier Robust Kalman Filter). We only implement filtering versions of these algorithms and not the smoothing since we are only concerned with the real-time solution in this work.

\subsection{Adaptive Loss Functions}

Similar to the Huber loss function described in HKF, other robust cost functions have been defined in statistics literature like the  Charbonnier, Cauchy, Geman-McClure, and Welsch losses. \cite{barron2019general} defines an adaptive version of the cost function which assumes the forms of these different loss functions based on the value of a tuning parameter $\alpha$. They optimize a modified version of this cost function with an added regularizer so that higher values of $\alpha$ de-weights outliers more but adds penalty for inliers, and lower values of $\alpha$ de-weights outliers less but also adds less penalty for inliers. Joint optimization in this filtering problem can be hard due to the high frequency of the IMU measurements, resulting in missed data. Similar to graduated non-convexity method discussed in \cite{yang2020graduated}, we can solve equation (\ref{eq:16}) iteratively with annealing the parameter $\alpha$ after a fixed number of iterations. We leave this for future work. 

\section{Experimental setup}
 The field tests for the method evaluation is done with a testbed rover (Pathfinder) (see Fig.~\ref{fig:pathfinder}). 
 The IMU, wheel encoder, and GPS receiver used in the rover, as similar to the work in \cite{kilic2019improved, kilic2021slip}, are provided in Table~\ref{tab:settings}. 
 To determine the reference truth solution, differential GPS (DGPS) is used. This solution is obtained by using two dual-frequency GPS receivers and Pinwheel (L1/L2)~\cite{novatel2} mounted both on a GPS Base Station and on the testbed rover. A loosely-coupled GPS/IMU fusion algorithm in \cite{kilic2019improved} is used for the state initialization process. The GPS measurements (carrier-phase and pseudorange) are recorded on both rover and base station receivers.  The reference truth solution is generated by post-processing on RTKLIB 2.4.2 software \cite{rtklib}. Note that, DGPS accuracy is expected as cm/dm level \cite{gps}. 
 
\begin{table}[h]
\centering
\caption{Pathfinder's Sensor Specifications}
\begin{tabular}{ l c c c c }
\hline
Sensor & & Model & Data Rate  \\

  \hline\hline
 IMU & & ADIS-16495~\cite{adis}  & 50 Hz \\  
 Wheel Encoder   & & Custom &  10 Hz  \\
 GPS Receiver &  & Novatel\cite{novatel1} & 10 Hz \\
 \hline
\end{tabular}

\label{tab:settings}
\end{table}


\begin{figure} [h!]
     \centering
     \begin{subfigure}[b]{0.3\columnwidth}
         \centering
         \includegraphics[width=1.25\textwidth]{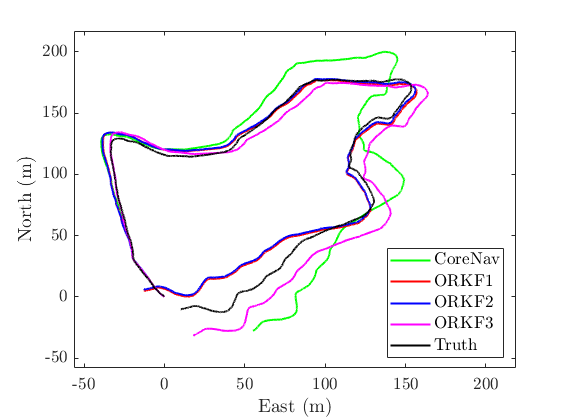}
         \caption{Ground track of $test1$}
         \label{fig:t9gt}
     \end{subfigure}
     \hfill
     \begin{subfigure}[b]{0.3\columnwidth}
         \centering
         \includegraphics[width=1.25\textwidth]{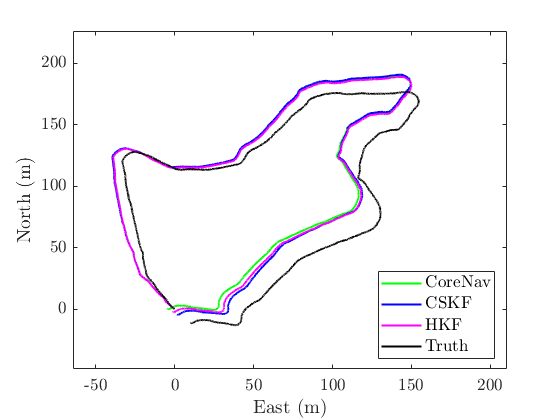}
         \caption{Ground track of $test2$}
         \label{fig:t11gt}
     \end{subfigure}
     \hfill
     \begin{subfigure}[b]{0.3\columnwidth}
         \centering
         \includegraphics[width=1.25\textwidth]{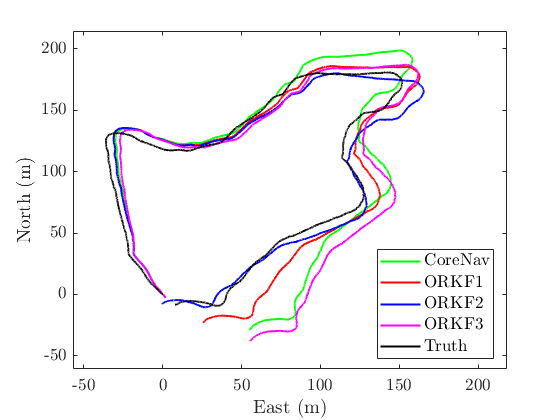}
         \caption{Ground track of $test3$}
         \label{fig:t11gt}
     \end{subfigure}
     \captionsetup{justification=centering}
        \caption{Ground Track (East-North planar trajectory) solutions of algorithms for data sets $test1$, $test2$, $test3$ \cite{vz7z-jc84-20}. For ease of display, we only show the trajectories that perform better than $CoreNav$ in each data set. Error statistics for all methods can be seen in Figures \ref{fig2:norm error_stats}, \ref{fig3:up error_stats}, Tables \ref{tab:1} and \ref{tab:2}.  }
        \label{fig1:ground track}
\end{figure}

To compare the above presented methods, we build upon the implementation in \cite{kilic2019improved} which uses Robot Operating System (ROS) \cite{quigley2009ros} for collecting data from sensors. As discussed in section \rom{3}, the base algorithm is an error-state EKF which uses IMU measurements (50 Hz) for state propagation and wheel odometry measurements (coming in at 10 Hz) as corrections. The state propagation step remains the same as $CoreNav$, whereas the correction step is modified to implement the robust methods. The HKF method depends on the parameter $\Delta$. This parameter is determined by the degree of contamination in the assumed Gaussian error distribution, and empirically tuned, in that it is difficult to estimate for the application of four wheel odometery. In this paper we selected that to be $1.5$. The CSKF method does not require any tuning parameters, and the critical value of Chi-square distribution of degree of freedom $4$ (size of measurement vector) with significance level of $0.05$, which is $9.488$ is used for outlier detection. In $ORKF1$, we select $s$ to $250$. $\nu$ is set to $300$ in $ORKF2$. Following \cite{van2004sigma}, the sigma-point calculations in (\ref{eq:29}) sets $\alpha$ to $1$, $\beta$ to $2$, and $\kappa$ to $0$. Following notations in \cite{yang2001adaptively}(referred here as $ORKF3$), $u$ is initialised to $2000$, $\tau$ also $2000$ with $\rho$ set to $0.9999$. All variational methods were run for $5$ iterations. This low number is partly due to the fact that increasing number of iterations resulted in skipping IMU measurements which is received at a very high rate within the real-time ROS implementation. Skipped IMU measurements, in turn, resulted unreliable results when repeated with the same data. 

For all implemented methods, parameters have been tuned for one of the data sets and re-used in the other ones, due to the similarity in the terrain and their trajectory shapes. The prior measurement and process noise remains the same for all methods. We talk about the data sets in the next section. They are also expected to vary from the ones mentioned above with different kinds of terrain requiring tuning from scratch. These parameters specify certain properties about the underlying process and measurement noise distribution which is discussed in the next section. In future work, we plan to learn these parameters from data acquired as the rover moves. 

\begin{figure*} [htb]
     \centering
     \begin{subfigure}[b]{0.3\columnwidth}
         \centering
         \includegraphics[width=1.1\textwidth]{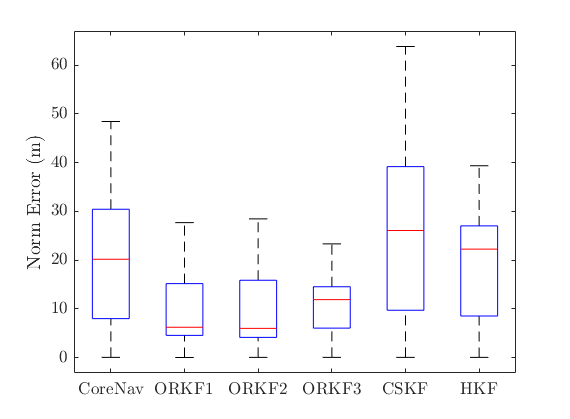}
         \caption{Norm Error (m) statistics of $test1$}
         \label{fig:t9n}
     \end{subfigure}
     \hfill
     \begin{subfigure}[b]{0.3\columnwidth}
         \centering
         \includegraphics[width=1.1\textwidth]{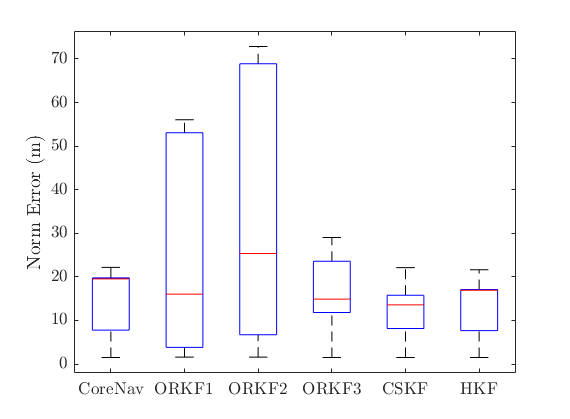}
         \caption{Norm error (m) statistics of $test2$}
         \label{fig:t9z}
     \end{subfigure}
     \hfill
     \begin{subfigure}[b]{0.3\columnwidth}
         \centering
         \includegraphics[width=1.1\textwidth]{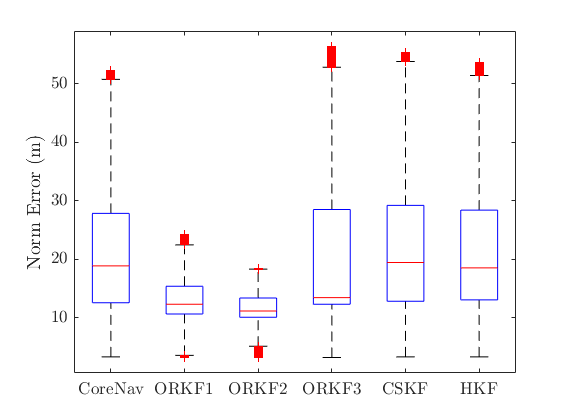}
         \caption{Norm error (m) statistics for $test3$}
         \label{fig:t11n}
     \end{subfigure}
         \caption{$L_2$ norm error statistics of all solutions for data sets $test1$, $test2$, $test3$}
        \label{fig2:norm error_stats}
\end{figure*}
\begin{figure*} [h!]
     \centering
     \begin{subfigure}[b]{0.3\columnwidth}
         \centering
         \includegraphics[width=1.1\textwidth]{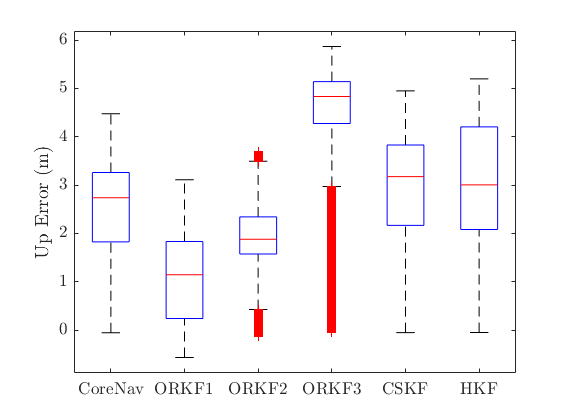}
         \caption{Up error (m) statistics of $test1$}
         \label{fig:t9n}
     \end{subfigure}
     \hfill
     \begin{subfigure}[b]{0.3\columnwidth}
         \centering
         \includegraphics[width=1.1\textwidth]{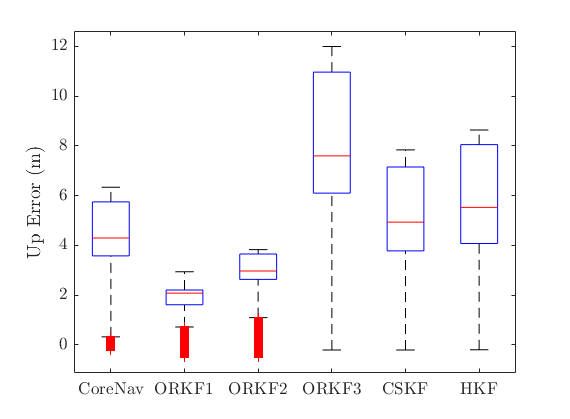}
         \caption{Up error (m) statistics of $test2$}
         \label{fig:t9z}
     \end{subfigure}
     \hfill
     \begin{subfigure}[b]{0.3\columnwidth}
         \centering
         \includegraphics[width=1.1\textwidth]{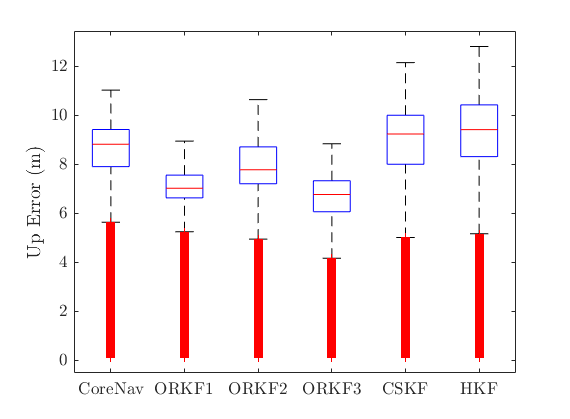}
         \caption{Up error (m) statistics of $test3$}
         \label{fig:t11n}
     \end{subfigure}
         \caption{Up error statistics of all solutions for data sets $test1$, $test2$, $test3$}
        \label{fig3:up error_stats}
\end{figure*}


\begin{figure*} [h]
     \centering
     \begin{subfigure}[b]{0.47\columnwidth}
         \centering
         \captionsetup{justification=centering,margin=1cm}
         \includegraphics[width=0.8\textwidth]{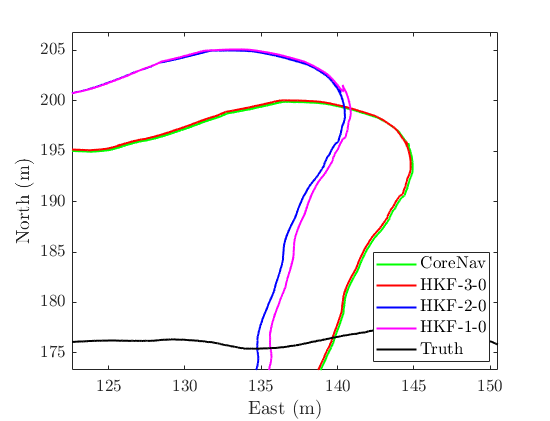}
         \caption{A magnified section of  $CoreNav$, $HKF$, and truth (GPS) trajectory for $test1$ dataset}
         \label{fig:t9n}
     \end{subfigure}
     \hfill
     \begin{subfigure}[b]{0.47\columnwidth}
         \centering
         \captionsetup{justification=centering,margin=1cm}
         \includegraphics[width=0.803\textwidth]{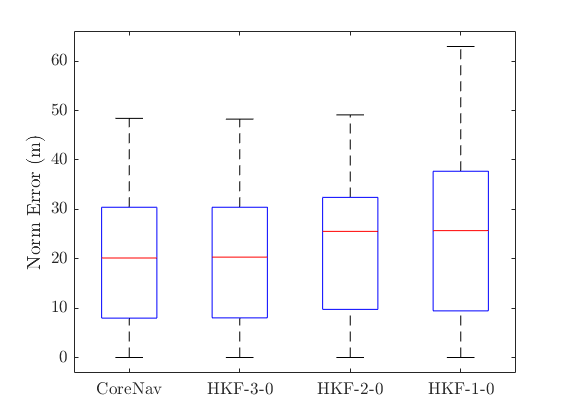}
         \caption{Norm Error (m) statistics of $CoreNav$ and $HKF$ solutions for $test1$ dataset}
         \label{fig:t9z}
     \end{subfigure}
     \captionsetup{justification=centering}
         \caption{Convergence of $HKF$ solutions to $CoreNav$ solutions with increasing $\Delta$ values from $1.0$ to $3.0$ for data set $test1$}
        \label{fig4:comparison HKF-CoreNav}
\end{figure*}
\section{Results and discussion}

The algorithms mentioned in section \rom{4} including $CoreNav$, ($HKF$, $CSKF$, $ORKF1$, $ORKF2$, $ORKF3$) are tested on the Pathtfinder platform (Fig. \ref{fig:pathfinder}) in three separate runs around a field of coal-ash residuals in Point Marion, PA. We use three data sets from \cite{vz7z-jc84-20} referred in the paper as $test1$ (\texttt{ashpile\_mars\_analog1.zip}), $test2$ (\texttt{ashpile\_mars\_analog2.zip}) and $test3$ (\texttt{ashpile\_mars\_analog3.zip}). This terrain was chosen to be reasonably representative of planetary terrains. The solutions of the algorithms and truth (DGPS) are shown in $ENU \, (East-North-Up)$ frame, which is the local navigation frame with origin at the start location for each the data sets. Figure \ref{fig1:ground track} shows the trajectories in the East-North plane. For ease of differentiating the trajectories, the figure shows only the methods that perform better compared to $CoreNav$ for each of the data sets. The 3D norm error and $Up$ error statistics are shown as box plots in Figures \ref{fig2:norm error_stats} and \ref{fig3:up error_stats} respectively. 

In data sets $test1$ and $test3$, $CSKF$ and $HKF$ have comparable performance to $CoreNav$ and $ORKF1$ and $ORKF2$ outperforms all other methods. In $test2$, however, the variational methods perform much worse with $ORKF1$ and $ORKF2$ being the worst affected. $CSKF$ and $HKF$ produce good results, better than $CoreNav$ in this case. We see a general improvement in $Up$ (vertical) axis solution with the variational filters $ORKF1$ and $ORKF2$.


\begin{table}[h!]
\centering
\begin{threeparttable}
\captionsetup{justification=centering}
\caption{RMS Errors (m) of all solutions for data sets $test1$, $test2$, $test3$ }
\begin{tabular}{ l c c c c }
\hline
 Method & & \(test1\) & \(test2\)  &  \(test3\) \\

  \hline\hline
 \(CoreNav\) & & 26.65 & 16.09 & 25.82\\  
 \(HKF\)   & & 21.78 &  \underline{14.67}  & 26.30\\
 \(CSKF\) &  & 34.76 & \textbf{13.82} & 27.18\\
 \(ORKF1\) &  & \underline{13.78} & 33.32 & \underline{14.08}\\
 \(ORKF2\) & & 14.13 & 43.86 & \textbf{11.67}\\
 \(ORKF3\) & & \textbf{13.34} & 18.18 & 26.13\\
 \hline
\end{tabular}

\begin{tablenotes}
\item{*} Best performance in bold and second-best underlined.
\end{tablenotes}

\label{tab:1}
\end{threeparttable}
\end{table}

\begin{table}[h!]
\centering
\begin{threeparttable}
\captionsetup{justification=centering}
\caption{Max. 3D Norm Errors (m) of all solutions for data sets $test1$, $test2$, $test3$ }
\begin{tabular}{ l c c c c }
\hline
 Method & & \(test1\) & \(test2\)  &  \(test3\) \\
 
 \hline \hline
 \(CoreNav\) & & 48.40 & 22.17 & 52.20\\  
 \(HKF\)   & & 39.34 &  \textbf{21.62}  & 53.57\\
 \(CSKF\) &  & 63.79 & \underline{22.13} & 55.40\\
 \(ORKF1\) &  & \underline{27.67} & 56.05 & \underline{24.15}\\
 \(ORKF2\) & & 28.43 & 72.85 & \textbf{18.33}\\
 \(ORKF3\) & & \textbf{23.31} & 29.04 & 56.35\\
 \hline
\end{tabular}
\begin{tablenotes}
\item{} 
\item{*} Best performance in bold and second-best underlined.
\end{tablenotes}
\label{tab:2}
\end{threeparttable}
\end{table}
It is important to understand what the parameters in each of these methods mean and their assumptions. In HKF, as discussed before the $\Delta$ parameter in (\ref{eq:12}) depends on a perturbing parameter $\epsilon$ which represents the amount of contamination in the Gaussian error distribution. For $\Delta = 0$, the error distribution is assumed to follow the Laplace distribution \cite{walck2007hand} and $\Delta \rightarrow \infty$ follows the Gaussian distribution. Thus the distribution of the innovations i.e ($y_{ik} - b_{ik}\hat{x}_{k/k}$) values in (\ref{eq:15}) will determine the best value of $\Delta$ at every time step. But with the less number of these innovations in each time step ($15$ states + $4$ measurements = $19$), it is hard to do so in this problem. However we do show experimentally that increasing $\Delta$ results in convergence with $CoreNav$ solution which assumes Gaussian error distribution (Figure \ref{fig4:comparison HKF-CoreNav}). This suggests all residuals in $HKF$ with $\Delta = 3.0$ are bounded by the nominal $R$ covariance ellipse.

CSKF assumes the large innovation is due to outliers in the measurements but could also be caused by wrong estimate of the predicted state. In that case, scaling the covariance matrix does not solve the problem. As discussed in \cite{chang2014robust}, another disadvantage is that an outlier in one dimension of a measurement vector results in the de-weighting of the whole vector. 

Variational filters also de-weight the measurements assuming heavy-tailed model which appears to be more effective in this case. Large value of $s$ ($250)$ in $ORKF1$ points to more importance given to the nominal measurement noise covariance. This is confirmed by the parameter value of $\nu$ ($300$) in $ORKF2$ which is close to the value of $s$ in $ORKF1$ since they essentially represent the same measurement model. Also the value of the $\lambda$ when running the algorithm remains within the range ($0.9,1$), which confirms our high belief in the nominal $R$ value. Similarly in $ORKF3$, $\rho$ value close to $1.0$ results in lesser effect of new residuals on the measurement noise estimate. The advantage of $ORKF3$ is the ability to set the degree of belief in the prior state uncertainty for every epoch. The large value of $\tau$ represents higher influence of the state uncertainty of the previous epoch on that of the current epoch. This can be favourable if the state uncertainty of previous epoch was  estimated correctly, but can get worse it is not the case. Overall, we see $ORKF3$ performs more consistently than the other two variational filters over the three datasets. The anomaly in the results is $test2$, which in spite of being very similar to $test1$ and $test3$, causes  $ORKF1$ and $ORKF2$ to perform poorly. One of the probable reasons could be the lack of convergence in the variational update step.
One of the main problems with applying the methods discussed above is the lesser number of measurements in this scenario unlike GPS or visual odometry which results in less information about the underlying measurement noise distribution. 

As mentioned before, the parameter $s$ in $ORKF1$ denotes the relative importance between the prior noise and sufficient statistics. There is another way of interpreting the effect of $s$. Looking at equation(\ref{eq:25}), the harmonic mean of the posterior distribution of covariance matrix is 

\begin{equation*}
    \Lambda_k = \frac{sR + S_k}{s+1}
\end{equation*}

which can be rewritten as 

\begin{equation*}
    \Lambda_k = \frac{sR}{s+1} + \frac{S_k}{s+1} .
\end{equation*}

The first part of the left hand side expression is the prior noise which is not affected by the measurements and also relatively unaffected by $s$ due to its ratio form. $S_k$ in the second part is large for outliers and smaller for inliers. Now considering only the case of outleirs, larger values of $s$ will produce a smaller $\Lambda_k$ than smaller values of $s$, thus de-weighting outliers less. Similar conclusion can be drawn for inliers as well. Thus lesser value of $s$ leads to more de-weighting of both inliers and outliers and vice versa. Also since $s$ is the degrees of freedom of the Student-t distrbuted measurement model in \cite{agamennoni2012approximate}, decreasing value of $s$ makes the distribution more heavy-tailed and thus greater weighting. Table (\ref{tab:3}) shows better localization performance for larger values of $s$, which suggests the true residual distribution is close to a Gaussian. 

\begin{table}[h!]
\centering
\begin{threeparttable}
\captionsetup{justification=centering}
\caption{RMS Errors (m) and Max. Norm Errors (m) of $ORKF1$ for data set $test1$ with varying values of $s$ }
\begin{tabular}{ l c c c}
\hline
 Method - $s$ & & $Max.\,Norm$(m) & $RMS(m)$ \\

  \hline\hline
 \(ORKF1-10\) & & 66.11 & 46.35 \\  
 \(ORKF1-50\)   & & \textbf{28.15} &  16.94 \\
 \(ORKF1-250\) &  & 29.20 & \textbf{14.45} \\
 \hline
\end{tabular}

\begin{tablenotes}
\item{} 
\item{*} Best performance in bold.
\end{tablenotes}

\label{tab:3}
\end{threeparttable}
\end{table}

\begin{table}[h!]
\centering
\begin{threeparttable}
\captionsetup{justification=centering}
\caption{RMS Errors (m) and Max. Norm Errors (m) of $ORKF2$ for data set $test1$ with varying values of $\nu$ }
\begin{tabular}{ l c c c}
\hline
 Method - $\nu$ & & $Max.\,Norm(m)$ & $RMS(m)$ \\

  \hline\hline
 \(ORKF2-10\) & & 83.54 & 50.28 \\  
 \(ORKF2-100\) & & 57.51 & 33.28 \\
 \(ORKF2-300\) &  & \textbf{20.50} & \textbf{14.78} \\
 \hline
\end{tabular}

\begin{tablenotes}
\item{} 
\item{*} Best performance in bold.
\end{tablenotes}

\label{tab:4}
\end{threeparttable}
\end{table}

Table (\ref{tab:4}) shows better localization performance for larger values of $\nu$ which agrees with results of $ORKF1$. This can explain why Huber cost function is not able to improve performance in this scenario. Since the residual distribution has been found to be close to Gaussian, the contamination parameter $\epsilon$ will be close to zero. \cite{karlgaard2010robust} shows the variation of optimal parameter in the Huber cost function  with respect to $\epsilon$, confirming $\Delta$ values greater than $2$ for $\epsilon$ close to zero. Thus greater values of $\Delta$ produced better results for $HKF$, since it agrees more with the true measurement noise distribution.




\section{Conclusion and future work}\label{ccl}
We extend upon the EKF in \cite{kilic2019improved} and implement robust filtering algorithms to increase localization accuracy in presence of erroneous wheel odometry measurements due to wheel slip. These methods are evaluated on a rover and seen to generally have improved performance over the standard EKF. The variational filters have better accuracy than others in $test1$ and $test3$. The covariance scaling and Huber regression methods show improvements in $test2$. Results point to a residual distribution that is close to a Gaussian distribution. Future work will include testing with more data sets over longer distances and online parameter learning. Another promising direction of work could be the use of adaptive robust cost functions which adapt to the type of terrain or slope the rover is traversing.  


\section*{Acknowledgment}
This work was supported in part by NASA EPSCoR Research Cooperative Agreement WV-80NSSC17M0053.
\bibliographystyle{IEEEtran}
\bibliography{references}
\end{document}